\documentclass[10pt,journal,compsoc]{IEEEtran}
\usepackage{amsmath}       
\usepackage{amssymb}       
\usepackage{amsfonts}      
\usepackage{graphicx}      
\usepackage{hyperref}      
\usepackage{cite}          
\usepackage{multirow}      
\usepackage{booktabs}     
\usepackage{algorithm}      
\usepackage{algorithmic}    
\usepackage{color}         
\usepackage{subcaption}    
\usepackage{textcomp}      
\usepackage{url}           
\usepackage{verbatim}     
\usepackage{array}        
\usepackage{graphicx} 
\usepackage{pifont}     
\usepackage{tabularx}   
\def\x{{\bf x}}

\def\I{{\cal I}}
\def\K{{\cal K}}

\def\P{{\bf P}}

\begin{document}

\title{Multi-label Classification with Panoptic Context Aggregation Networks}

\author{Mingyuan Jiu, Hailong Zhu, Wenchuan Wei, Hichem Sahbi, Rongrong Ji, Mingliang Xu

\thanks{M.~Jiu, H.~Zhu, W.~Wei and M.~Xu are with School of Computer and Artificial Intelligence, Zhengzhou University, Zhengzhou, China, Engineering Research Center of Intelligent Swarm Systems, Ministry of Education, China and National Supercomputing Center in Zhengzhou, Zhengzhou, China. H.~Sahbi is with Sorbonne University, CNRS, LIP6, F-75005, Paris, France. R.~Ji is with Xiamen University, Xiamen, China.} 
\thanks{This work is supported by the National Natural Science Foundation of China (No.s~62272422, U22B2051), and also partially by the Natural Science Foundation of Henan Province (No.~252300421225) and Organized Young Scientific Research Team Cultivation Foundation of Zhengzhou University (No.~35220549).} 
}

\maketitle

\begin{abstract}
  Context modeling is crucial for visual recognition, enabling highly discriminative image representations by integrating both intrinsic and extrinsic relationships between objects and labels in images. A limitation in current approaches is their focus on basic geometric relationships or localized features, often neglecting cross-scale contextual interactions between objects. This paper introduces the Deep Panoptic Context Aggregation Network (PanCAN), a novel approach that hierarchically integrates multi-order geometric contexts through cross-scale feature aggregation in a high-dimensional Hilbert space. Specifically, PanCAN learns multi-order neighborhood relationships at each scale by combining random walks with an attention mechanism. Modules from different scales are cascaded, where salient anchors at a finer scale are selected and their neighborhood features are dynamically fused via attention. This enables effective cross-scale modeling that significantly enhances complex scene understanding by combining multi-order and cross-scale context-aware features. Extensive multi-label classification experiments on NUS-WIDE, PASCAL VOC2007, and MS-COCO benchmarks demonstrate that PanCAN consistently achieves competitive results, outperforming state-of-the-art techniques in both quantitative and qualitative evaluations, thereby substantially improving multi-label classification performance.
\end{abstract}

\begin{IEEEkeywords}
Context-aware kernel networks, cross-scale \& multi-order representations, neural networks, multi-label classification
\end{IEEEkeywords}

\section{Introduction} \label{sec;intro}

\IEEEPARstart{T}{he} field of multi-label image classification~\cite{tsoumakas2007multi} stands as a cornerstone in contemporary computer vision, driven by its pervasive utility across diverse real-world applications. Within this paradigm, an image is not merely assigned a single identifier but rather a rich set of semantic labels. Crucially, the intricate relationships and interdependencies among these co-occurring labels encapsulate vital contextual information regarding the visual content~\cite{wang2016cnn}. This inherent contextual knowledge, particularly the relationships between depicted patterns, is hypothesized to significantly influence the predictive accuracy of machine learning models. Therefore, the effective exploitation of such contextual cues is paramount; it offers crucial insights that empower advanced algorithms to achieve a more profound understanding of image content, ultimately leading to a substantial enhancement in classification performance.

Despite the remarkable progress of deep learning in multi-label classification, many current models primarily emphasize localized appearance features. This reliance often hinders their capacity to effectively discern contextual dependencies~\cite{wang2016cnn}, thereby restricting their ability to accurately identify objects that exhibit significant scale variations, occlusions, or intricate spatial arrangements across diverse scales~\cite{zhao2024multi}. For instance, consider an image containing multiple objects: when confronted with substantial disparities in object size, severe occlusion, or complex background clutter~\cite{xiao2020noise}, cross-scale contextual information becomes crucial for robust detection. A pedestrian on a street, for example, typically necessitates a smaller, more localized context, whereas a vehicle often requires broader context.

To address cross-scale context-aware modeling problem, there are many works emphasizing the integration of both local details and global semantic structures \cite{zhou2024mscanet,xu2023progressive,zhang2021global}. Cross-scale context analysis not only captures geometric details but also incorporates global semantic relationships, ensuring to maintain consistent feature representations when handling objects at different scales~\cite{li2024enhanced}. Recent advancements, such as hierarchical attention-based context-aware networks~\cite{liu2024vision,ji2021hierarchical}, dynamically adjust the model's focus on different scale regions, significantly improving model performance in handling objects of various sizes and also occlusion~\cite{ren2023multi}. Additionally, emerging techniques like Graph Neural Networks (GNNs) have also been applied to multi-label classification, further optimizing the use of contextual information~\cite{singh2022multi}. These methods have demonstrated strong performance in both semantic segmentation and multi-label classification, especially in complex scenes where they are able to enhance the robustness and accuracy of models.

This paper introduces a novel deep Panoptic Context Aggregation Network (PanCAN), building upon the foundational principles of context-dependent kernels~\cite{Sahbicvpr2008, sahbi2015imageclef, JiuSahbiicpr2018, JiuSahbi2022nc}. We reformulate the kernel mapping solution, derived from the optimization of an objective function, as a deep neural network, where each layer corresponds to a single iteration of this optimization. Within each layer, we aggregate cross-scale geometric and semantic information to yield a comprehensive set of context features. In practice, images are partitioned into a lattice of cells at various contexts with each context proportionally defined by the dimension of the underlying cell. By fusing multiple scale contexts—ranging from fine-grained to coarse-grained—within a single layer, our model effectively captures visual and structural information spanning micro-and-macro contexts, significantly enhancing classification performance in complex scenes. This approach extends our preliminary work in~\cite{JiuZhuicpr2024} by incorporating cross-scale contexts, ranging from micro-patches to macro-patches, for improved representation, which leads to an extra gain in performance. This is further substantiated by a more comprehensive analysis and extensive experiments across diverse benchmarks, including NUS-WIDE, PASCAL VOC2007, and MS-COCO. \\ Considering the aforementioned challenges, the primary contributions of this paper include 
\begin{itemize}
\item A multi-order context-aware kernel mapping that addresses the limitations of coarse-grained contexts in existing kernel networks, learning fine-grained and extended-range contexts across different extents by incorporating random walk and attention mechanisms. 
\item To account for varying contextual scales, cross-scale context-aware kernel mapping is further proposed. This integrates contexts of diverse scales, progressively constructing macro-contexts hierarchically from micro-contexts.
\item Extensive experiments on three publicly available benchmark datasets show that our method surpasses state-of-the-art models. The results show that the proposed framework significantly enhances model robustness and accuracy when handling complex and diverse scenes.
\end{itemize} 
The rest of this paper is organized as follows: we firstly review the related works in Section~\ref{sec:relatedwork}, and then we revisit our previous context-aware kernel design and its learning algorithm~\cite{JiuSahbi2022nc} in Section~\ref{sec:contextaware}. In Section~\ref{sec:framework}, we show our two main contributions about deep Panoptic Context Aggregation Network and its end-to-end learning algorithm. In Section~\ref{sec:experiments}, we show the comparison performance and make comprehensive analysis for the proposed method on the NUS-WIDE, PASCAL VOC2007 and MS-COCO benchmarks. Finally, we conclude the paper in Section~\ref{sec:conclusion} and provide possible extensions for a future work.

\section{Related Work} \label{sec:relatedwork}

\subsection{Multi-label classification}

Multi-label classification, a core task in computer vision, involves simultaneously describing each image with multiple semantic labels~\cite{zhu2023scene}. While most existing methods primarily focus on natural images with common object labels like ``bicycle'',``table'', etc., early research endeavored to model label co-occurrence from single visual inputs. Techniques such as Recurrent Neural Networks (RNNs) and graph-based models were employed to explore semantic relationships between labels~\cite{wang2016cnn}, with Long Short-Term Memory networks (LSTMs)~\cite{hua2019deep} also utilized to learn dependencies across distinct semantic regions. Other works adopted dynamic Graph Convolutional Networks (GCNs)~\cite{ye2020attention,manessi2020dynamic} to model context-aware relationships among co-occurring categories.

More recently, contextual and cross-scale analysis have gained prominence in learning implicit relationships. Transformer models~\cite{pereira2024review} are now applied to explore connections between visual features and labels, and self-attention mechanisms~\cite{zhu2020deformable} are used to capture spatial relationships between objects. These advancements significantly improve model performance by enhancing the understanding of complex contexts. Further methods, such as ML-Decoder~\cite{ridnik2023ml} and Q2L~\cite{liu2021query2label}, leverage multi-attention layers or Transformer decoders to boost performance. However, relying solely on local visual features are insufficient to represent complex image content. Since objects exist through multiple scales within an image, models dependent on local features struggle to capture cross-scale interactions. Consequently, recent approaches have shifted towards cross-scale context-aware modeling~\cite{ji2021multi,rangarajan2022ultra}, enabling to dynamically learn feature representations. This cross-scale context representation allows models to retain fine-grained local details while incorporating broader semantic information, thereby improving the accuracy of multi-label classification.

Our proposed framework is designed to integrate both cross-scale and multi-order contexts for multi-label classification. It encodes visual features by aggregating associated contextual information, \textcolor{black}{not only to capture intricate local details but also to model complex interactions between parts in a given object or between objects through contexts at different scales}. Unlike traditional attention-based models that primarily focus on assigning varying weights within specific regions, often overlooking inter-scale contextual information, our approach integrates features from diverse scales by constructing a set of dynamic anchors. This fully leverages both micro- and macro-contextual information in complex scenes, ultimately enhancing classification performance.

\subsection{Context modeling}

Context modeling is an important technique in computer vision and natural language processing. It aims to enhance a model's understanding of complex scenes by capturing the relationships between scene patterns~\cite{nagaraja2016modeling}. Early context modeling approaches often relied on designed local features (e.g., HOG~\cite{DalalTriggsCVPR05}). However, they were inadequate to learn optimal global relationships within images. The introduction of the self-attention mechanism brought groundbreaking advancements in context modeling. The self-attention mechanism in Transformers~\cite{li2021local} enables the establishment of global dependencies between different positions, allowing the model to capture complex spatial relationships between objects. Graph Convolutional Networks (GCNs) have also been applied to context modeling, where image or text data are represented as graph structures, and GCNs learn the relationships between objects through nodes and edges in graphs~\cite{kipf2016semi}.

Recent advancements in context modeling have demonstrated significant progress; by leveraging cross-scale analysis, the models can jointly preserve fine-grained local details and high-level semantic structures. Encoder-decoder architectures~\cite{ates2023dual} extract features at various scales, capturing multiple layers of contextual information, which has led to significant progress in fields like medical image segmentation~\cite{zhu2024medical}. Feature Pyramid Networks (FPNs)~\cite{lin2017feature} combined feature maps at different resolutions (scales) to enhance contextual information within images, allowing the model to better handle variations in object size and complex backgrounds.

Our proposed methodology advances the context-aware kernel network through an objective function, designed to yield an optimized kernel mapping via multi-order and cross-scale context aggregation. Central to this design, a dynamic multi-order contextual scheme is employed to delineate both fine-grained local and coarse-grained global contextual information. Moreover, the collaborative integration of random walk and attention mechanisms refines the derived context structure: the attention mechanism precisely directs the model's focus toward salient regions within designated anchor cells, while the random walk likewise facilitates the exploration of optimal higher-order relational configurations among these cells. Additionally, a dynamic cross-scale context aggregation module enables the generation of robust feature representations.

\subsection{Cross-scale modeling}

Different strategies are widely employed to enhance model discriminative power through the extraction of features and structures at various resolutions, e.g., Feature Pyramid Networks~\cite{lin2017feature}. This approach enables precise detection of objects across diverse scales and significantly enhances model robustness. Attention mechanisms are also integrated within Transformer-based architectures, such as Swin Transformer~\cite{liu2021swin} and SegFormer~\cite{xie2021segformer}, enabling Transformer layers to attend to information at diverse scales. The attention network in ~\cite{wang2024multi} further synergistically combines the recently proposed large-kernel attention mechanism with a cross-scale approach to generate attention maps at varying levels of granularity. Spatial attention has also been proposed to extract contextual information, demonstrating excellent performance in tasks such as super-resolution~\cite{zhao2019compression}. Beyond computer vision, cross-scale modeling finds applications in other fields, notably remote sensing, where MSCANet~\cite{zhou2024mscanet} integrates cross-scale deep and shallow feature maps for detecting small objects within large remote sensing images. Collectively, these models are capable of combining both fine-grained local and coarse-grained global features, resulting in significant performance gains.

A panoptic strategy is adopted in the proposed network. Rather than employing a parallel architecture for multiple scales, features are built in a cascaded fashion, progressing from fine (emphasizing detailed information) to coarse (capturing broader structural context). Specifically, the output of a finer-scale module serves as the input to a coarser module. Through end-to-end learning, the offsets between fine-and-coarse-scale patches are automatically learned within the network during image partitioning. This dynamic selection and aggregation collectively enable the capture of both local details and global structures across multiple scales.
\section{Revisit Context-aware Kernel Networks} \label{sec:contextaware}

In this section, we revisit the principle of {\it context-aware kernel maps} and the underlying {\it context networks} that operate on images partitioned into regular grids of cells.
\subsection{Context-aware Kernel Maps}

Let $\{\mathcal{I}_p\}_{p=1}^P$ denote a set of labeled training images, where $Y_l^p$ is a binary label indicating the membership of image $\mathcal{I}_p$ to class $l \in \{1, \ldots, L\}$. For a given scale, the set $\mathcal{S}_p = \{\mathbf{x}_1^p, \ldots, \mathbf{x}_n^p\}$ represents non-overlapping cells sampled from a regular grid within the image $\mathcal{I}_p$. Without loss of generality, $n$ is assumed constant for a fixed scale. The objective of context-aware kernel design is to learn a function $\kappa$ that assigns higher similarity to cells that are not only visually but also geometrically similar. The resultant kernel matrix $\mathbf{K}$ is optimized via the following objective function~\cite{JiuSahbiicpr2018,sahbi2015imageclef}
\begin{equation} \label{eq:context-awareloss}
\min_{\mathbf{K}} \text{tr}(-\mathbf{K} \mathbf{S}^\top) - \alpha \sum_{c=1}^C \text{tr}(\mathbf{K} \mathbf{P}_c \mathbf{K}^\top \mathbf{P}_c^\top) + \frac{\beta}{2} \| \mathbf{K} \|_2^2,
\end{equation}
being $\mathbf{K}$  a gram matrix with $[\mathbf{K}]_{\mathbf{x}_i,\mathbf{x}_j} = \kappa(\mathbf{x}_i,\mathbf{x}_j)$ and $\mathbf{S}$ a context-agnostic matrix computed over the entire set of cells $\mathcal{X} = \bigcup_p \mathcal{S}_p$, whilst $\alpha \geq 0$ and $\beta > 0$ are hyperparameters controlling the trade-off between different terms. In the above objective function, $\text{tr}(\cdot)$ denotes the trace operator, and $\top$ indicates matrix transposition. The set $\{\mathbf{P}_c\}_{c=1}^C$ represents $C$ distinct types of predefined neighborhood relationships among cells (e.g., $C=4$ for up, down, left, and right spatial adjacencies, as in~\cite{JiuSahbi2022nc}). Specifically, for a given cell $\mathbf{x}$, if there exists a cell $\mathbf{x}'$ related by the $c$-th neighborhood type, then $[\mathbf{P}_{c}]_{\mathbf{x},\mathbf{x}'} \neq 0$; otherwise, $[\mathbf{P}_{c}]_{\mathbf{x},\mathbf{x}'} = 0$ for all $\mathbf{x}' \in \mathcal{X}$. In Eq.~\eqref{eq:context-awareloss}, the first term acts as a fidelity criterion, encouraging high similarity values for visually similar cell pairs $\{(\mathbf{x}_i, \mathbf{x}_j)\}_{ij}$. The second term serves as a context criterion, promoting high similarity between cell pairs exhibiting similar neighborhood relationships. The third term is a regularizer, ensuring a well-posed optimal solution.

One may show that the  solution of Eq.~\eqref{eq:context-awareloss} is given as the fixed-point of the following recursive function 
\begin{equation} \label{eq:3}
\mathbf{K}^{[\textrm{t+1}]} = \mathbf{S} + \gamma \sum_{c=1}^C \mathbf{P}_c \ \mathbf{K}^{[\textrm{t}]} \ \mathbf{P}_c^{\hspace{-0.05cm}\top},
\end{equation}
where the superscript $\textrm{t}$ refers to the iteration number, and $\gamma = \alpha / \beta$ ensures convergence to a stable solution. According to the Representer Theorem, Eq.~\eqref{eq:3} can be equivalently expressed as an explicit mapping within a high-dimensional Hilbert space
\begin{equation} \label{eq:4}
\mathbf{\Phi}^{[\textrm{t+1}]} = \left( {\mathbf{\Phi}^{[0]}}^{\top} \quad \gamma^{\frac{1}{2}} \mathbf{P}_1 {\mathbf{\Phi}^{[\textrm{t}]}}^{\top} \quad \ldots \quad \gamma^{\frac{1}{2}} \mathbf{P}_C {\mathbf{\Phi}^{[\textrm{t}]}}^{\top} \right)^{\top},
\end{equation}
here $\mathbf{\Phi}^{[\textrm{t}]}$ corresponds to the kernel mapping of $\mathbf{K}^{[\textrm{t}]}$, and $\mathbf{\Phi}^{[0]}$ represents either an exact or an approximate mapping of $\mathbf{S}$ for all $\mathbf{x} \in \mathcal{X}$.

\subsection{Context-aware Kernel Networks}

Given established neighborhood relationships among cells, Eq.~\eqref{eq:4} describes the iterative procedure for the context-aware kernel mapping. At each iteration, the value $[\mathbf{K}^{[\textrm{t}]}]_{\mathbf{x}_i,\mathbf{x}_j}$ between two cells $\mathbf{x}_i$ and $\mathbf{x}_j$ can be computed as the inner product of their corresponding unfolded kernel mappings
\begin{align} \label{eq:18}
    [\mathbf{K}^{[\textrm{t}]}]_{\mathbf{x}_i,\mathbf{x}_j} &= \mathbf{\phi}^{[\textrm{t}]} \big( \ldots \big(\mathbf{\phi}^{[1]}(\mathbf{\phi}^{[0]}(\mathbf{x}_i)) \big) \big) \cdot \nonumber \\
    & \quad \quad \quad \mathbf{\phi}^{[\textrm{t}]} \big(\ldots\big(\mathbf{\phi}^{[1]}(\mathbf{\phi}^{[0]}(\mathbf{x}_j))\big)\big),
\end{align}
where $\mathbf{\phi}^{[\textrm{t}]}(\cdot)$ represents a column vector of $\mathbf{\Phi}^{[\textrm{t}]}$. When the limit of the iteration number is set to $T$, the process outlined in Eq.~\eqref{eq:18} can be interpreted as a $T$-layer context-aware kernel network~\cite{JiuSahbiicpr2018,JiuSahbi2022nc} where each layer corresponds to a single iteration of Eq.~\eqref{eq:4}. In the input layer, $\mathbf{\phi}^{[0]}$ is initialized using a pre-trained visual model (e.g., ResNet101~\cite{he2016deep}). For a hidden layer $t$, the mapping $\mathbf{\phi}^{[t]}$ is updated using the adjacency matrices as per Eq.~\eqref{eq:4}. Finally, the output layer yields the approximate context-aware mapping $\mathbf{\phi}^{[T]}(\mathbf{x})$ for a given cell $\mathbf{x}$.

In order to measure the similarity between images $\mathcal{I}_p$ and $\mathcal{I}_q$, a convolution kernel---between their respective cells---is obtained  as
\begin{align} \label{eq:1}
   \mathbf{K}(\mathcal{I}_p, \mathcal{I}_q) &= \sum_{i, j} \kappa^{[\textrm{T}]}(\mathbf{x}_i^p, \mathbf{x}_j^q), \nonumber\\
   &= \left( \sum_{\mathbf{x}_i \in \mathcal{S}_p} \mathbf{\phi}^{[\textrm{T}]}(\mathbf{x}_i^p) \right) \cdot \left( \sum_{\mathbf{x}_j \in \mathcal{S}_q} \mathbf{\phi}^{[\textrm{T}]}(\mathbf{x}_j^q) \right) \nonumber \\
   &= \langle \mathbf{\phi}_{\mathcal{K}}(\mathcal{S}_p), \mathbf{\phi}_{\mathcal{K}}(\mathcal{S}_q) \rangle,
\end{align}
where $\kappa^{[\textrm{T}]}$ is the learned context-aware kernel evaluated on $\mathcal{I}_p \times \mathcal{I}_q$. The image features $\mathbf{\phi}_{\mathcal{K}}(\mathcal{I}_p)$ are thus obtained as
\begin{equation}\label{eq:img_feats}
    \mathbf{\phi}_{\mathcal{K}}(\mathcal{I}_p) = \sum_{i=1}^{n} \mathbf{\phi}^{[\textrm{T}]}(\mathbf{x}_i^p), \quad \forall \mathbf{x}_i^p \in \mathcal{S}_p.
\end{equation}
The network defined by Eq.~\eqref{eq:18} shares similarities with common deep learning architectures; however, its architecture is dynamically and automatically obtained as the solution of the objective function in Eq.~\eqref{eq:context-awareloss}. To fully exploit its potential, an end-to-end framework is adopted to learn (layer-wise) the neighborhood matrices $\{\mathbf{P}_c^{[t]}\}_c$ as discussed later.

\section{Proposed PanCAN} \label{sec:framework}
\noindent Extending the principles of the context-aware kernel network introduced in Section~\ref{sec:contextaware}, we herein present the Deep Panoptic Context Aggregation Network (PanCAN) for multi-label image classification. \textcolor{black}{Here the {\it contextual order} for a cell is defined by the range of its neighborhood, where higher orders denote larger contextual distances. The {\it scale} of a cell refers to its spatial resolution, with macro-scale cells representing larger regions and micro-scale cells representing smaller, finer-grained regions.} This novel  (PanCAN) network is characterized by several key properties: it is recursively built from micro- to macro-scale cells, where each macro-scale cell constitutes an aggregation of micro-cells, ensuring that learned features encapsulate both fine-grained local details and broad global semantic information. Furthermore, multi-order contexts are learned in each layer; in other words, the directional and structural relationships of neighboring cells are explicitly considered, enabling the capture of extended-range, structured context. Inspired by~\cite{iscen2022learning,iscen2019label}, the proposed network is also designed end-to-end, enabling hierarchical label propagation from macro- to micro-scale cells, a process that significantly enhances the overall performance of multi-label image classification.
\subsection{Multi-order Context-aware Mapping Network} \label{subsec:multiorder}

\begin{figure}[tbp]
    \centering 
    \includegraphics[width=0.6\linewidth]{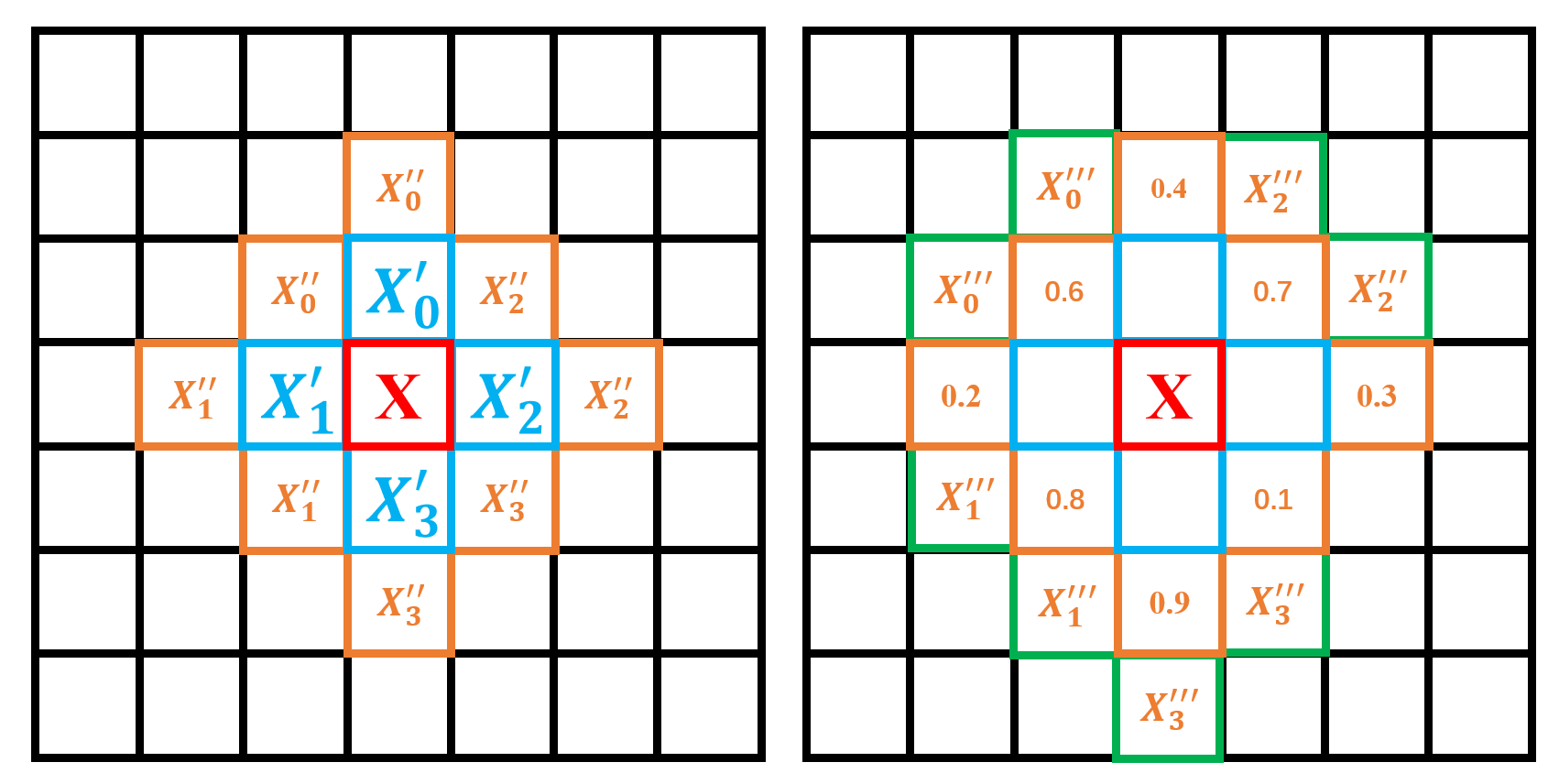}
    \caption{Multi-order neighborhood system. The left side shows the first-order (blue cells) and second-order neighborhoods (orange cells). On the right, the third-order neighborhood (green cells) is built from the second-order neighborhood based on the transition probabilities.} \label{fig:Domain_construction}
\end{figure}

In conventional context-aware kernel networks, a significant limitation is the {\it undifferentiated} contribution of all cells within a given $c$-typed neighborhood to context definition. This approach---which treats varying extents of proximity within the neighborhood equivalently---may constrain expressivity and can lead to an inaccurate or incomplete representation of the actual context. To overcome this issue, we introduce a multi-order neighborhood for any cell $\mathbf{x}$, as depicted in Fig.~\ref{fig:Domain_construction}-left. For a given cell $\mathbf{x}$, its first-order ($c$-typed) neighborhood, denoted $\{\mathcal{N}_c^{(1)}(\mathbf{x})\}_c$, is implicitly defined by the adjacency matrices $\{\mathbf{P}_c \}_{c=1}^C$, where $c$ indexes distinct neighborhood systems. Higher-order neighborhoods (for $k \geq 2$) are recursively generated as
\begin{equation} \label{eq:5}
{\cal N}_c^{(k)}(\mathbf{x})=\displaystyle \bigcup_{\mathbf{x}'\in {\cal N}_c^{(k-1)}(\mathbf{x})} {\cal N}_c^{(k-1)}(\mathbf{x}') \quad \text{with} \quad \mathbf{x}' \neq \mathbf{x}.
\end{equation}
This recursive formulation inherently captures increasingly long-range contextual information. To effectively leverage these higher-order relationships, we propose a combined random walk and attention mechanism applied to the $k$-th order neighborhood, yielding a structured graph. The core idea is to estimate similarity probabilities for random walks through an attention mechanism. Specifically, the similarity between a central cell $\mathbf{x}$ and any cell $\mathbf{x}''$ within its $k$-th order neighborhood $\mathcal{N}_c^{(k)}(\mathbf{x})$ is quantified by their attention score as
\begin{equation} \label{eq:6}
\text{score}(\mathbf{x}, \mathbf{x}'')  = \frac{\left(W_q^{k} \mathbf{\phi}(\mathbf{x})\right)^{\hspace{-0.05cm}\top} W_m^{k} \mathbf{\phi}(\mathbf{x}'')}{\sqrt{d}},
\end{equation}
where $\mathbf{\phi}(\cdot)$ denotes the feature vector of a cell, $W_q^{k}$ and $W_m^{k}$ are learnable parameter matrices specific to the $k$-th order, and $d$ \textcolor{black}{is the dimensionality of $\mathbf{\phi}(\cdot)$.} Subsequently, the conditional probabilities $f_c^{(k)}(\mathbf{x}'' | \mathbf{x})$ for random walks from $\mathbf{x}$ to its $k$-th order neighbors $\mathbf{x}'' \in \mathcal{N}_c^{(k)}(\mathbf{x})$ (for a given $c$) are obtained by applying a softmax normalization as
\begin{equation} \label{eq:7}
f_c^{(k)}(\mathbf{x}'' | \mathbf{x}) = \frac{\exp(\text{score}(\mathbf{\phi}(\mathbf{x}), \mathbf{\phi}(\mathbf{x}'')))}{\sum_{z\in \mathcal{N}_c^{(k)}(\mathbf{x})} \exp(\text{score}(\mathbf{\phi}(\mathbf{x}), \mathbf{\phi}(z)))}.
\end{equation}
\noindent Leveraging these transition probabilities, we refine the features for a cell $\mathbf{x}$ in each $c$-typed neighborhood by incorporating its $k$-th order context. This refinement is expressed as a weighted sum of the cell features within the $k$-th order neighborhood $\mathcal{N}_c^{(k)}(\mathbf{x})$, using their corresponding transition probabilities $\{f_c^{(k)}(\mathbf{x}''| \mathbf{x})\}_{\mathbf{x}''}$ as
\begin{equation} \label{eq:8}
  \mathbf{\phi}_{c,k}(\mathbf{x}) = \sum_{{ \mathbf{x}''\in {\cal N}_c^{(k)}(\mathbf{x}) }} f_c^{(k)}(\mathbf{x}''| \mathbf{x}) (W_v^{k}\mathbf{\phi}(\mathbf{x}'')),
\end{equation}
where $W_v^{k}$ denotes learnable parameters. The combination of Eqs~\eqref{eq:6}-\eqref{eq:8} thus defines the procedure for constructing $k$-th order context-aware features.

While this procedure, depicted in Fig.~\ref{fig:Domain_construction}-right, can be progressively extended to build features incorporating larger contexts, it is noteworthy that although higher-order neighborhoods capture extended-range dependencies, the semantic influence of individual cells within these broader contexts tends to diminish. Consequently, to mitigate this, our empirical approach, validated in the experimental section, considers only those cells exhibiting high transition probabilities $\{f_c^{(k)}(\mathbf{x}''| \mathbf{x})\}_{\mathbf{x''}}$, rather than including all neighboring cells, in the construction of higher-order context-aware features.
\begin{figure*}[tbp]
    \centering 
    \includegraphics[width=\textwidth]{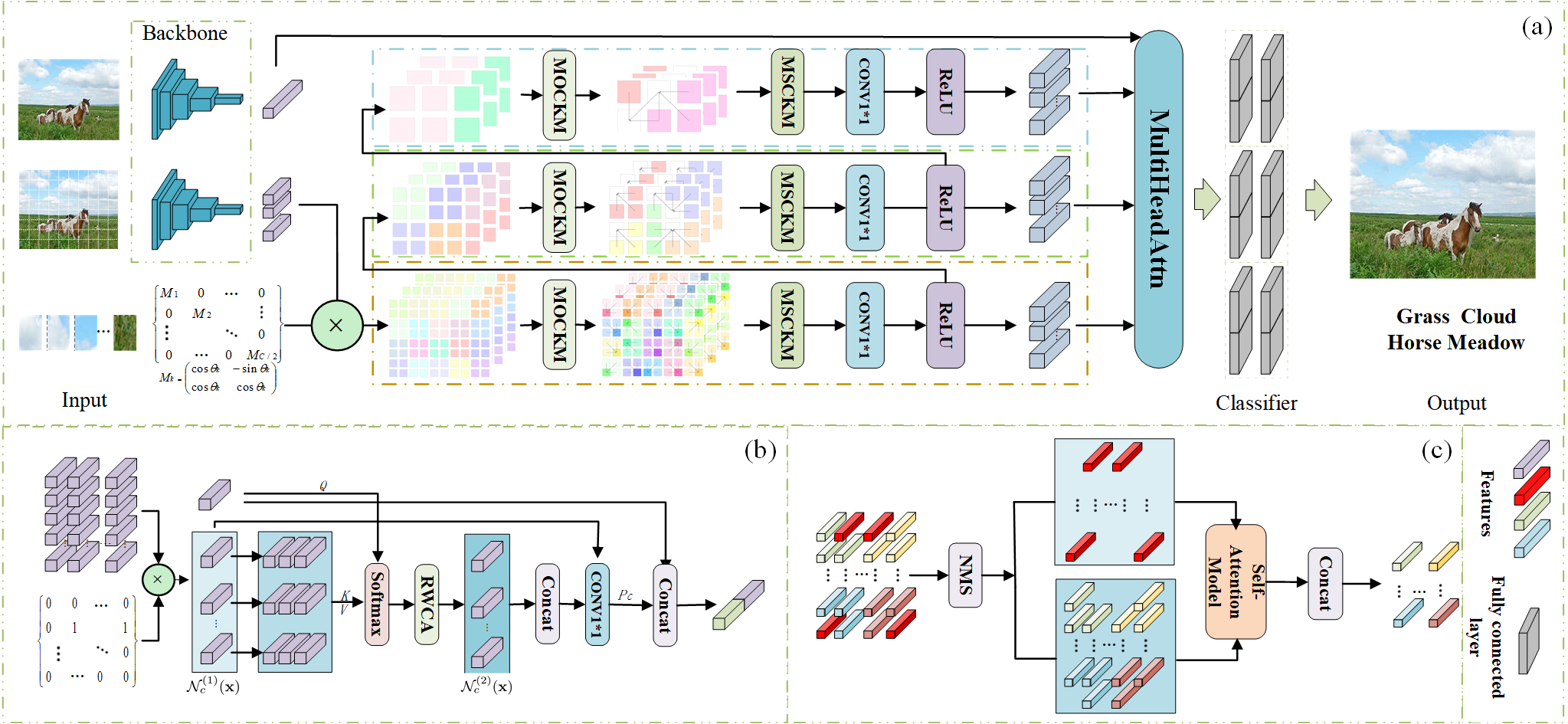} 
    \caption{The proposed PanCAN. (a) The overall architecture: the input image is divided into a grid of cells, where each cell is represented by visual features and rotated positional features. These enhanced features are then processed by PanCAN to capture fine-grained and cross-scale contexts, which are further fused with global features for classification. (b) ``Multi-order context-aware mapping network'' module (``MOCAMN''), where ``RWCA'' denotes the Random Walk-based Context Aggregation mechanism. The colored boxes represent different levels of neighborhood relations: first-order neighbors, directly adjacent blocks (i.e., neighbors of first-order blocks), and second-order neighbors. (c) ``Cross-scale context-aware mapping network'' module (``CSCAMN''), the most salient cell in red within each macro-cell is estimated as the anchor, while the remaining cells in different colors belonging to the same macro-cell are aggregated with the anchor through an attention mechanism to obtain the representation of macro-cells.} \label{fig:framework}
\end{figure*}

\subsection{Cross-scale Context-aware Mapping Network} \label{subsec:multiscale}

We introduce a cross-scale context-aware mapping network {\it (module)} that integrates information across diverse scales through attention-based fusion, capturing both localized and global contextual features. An image, characterized by its $H \times W$ dimensions, is hierarchically decomposed into $S$ distinct scales, denoted $\{R^{(s)}\}_{s=1}^S$. Each $R^{(s)}$ forms a grid of cells $\{\mathbf{x}_{i, j}^{(s)} \mid (i, j) \in (N_h, N_w)\}$, where $N_h = \lceil H/\Delta_h \rceil$, $N_w = \lceil W/\Delta_w \rceil$ and  $\lceil . \rceil$ stands for the ceiling function. Here, $\Delta_h$ and $\Delta_w$ define the vertical and horizontal overlap strides, respectively. The inter-scale relationship is defined by these strides, ensuring each coarser-grained macro-cell $\mathbf{x}_{i, j}^{(s)}$ encompasses multiple overlapping finer-grained micro-cells $\mathbf{x}_{i, j}^{(s-1)}$; more precisely, for a given macro-cell at scale $s$, its features are obtained by {\it fusing} the multi-order context-aware features derived from its constituent micro-cells. This fusion is weighted by similarities to an {\it anchor} micro-cell, $\mathbf{x}_a^{(s-1)}$. This anchor corresponds to the top-ranked micro-cell (within each macro-cell $\mathbf{x}^{(s)}$) according to the  $\ell_2$-norm of its context-aware features, and subsequently refined through non-maximum suppression to guarantee spatial diversity. 

\indent With the anchor $\mathbf{x}_a^{(s-1)}$ defined, the features of $\mathbf{x}^{(s)}$ are obtained by aggregating its constituent micro-cells  $\{\mathbf{x}^{(s-1)}_m\}_m$  via an attention mechanism, which captures long-range dependencies. The attention weights $f_m^{(s-1)}$ between the anchor $\mathbf{x}_a^{(s-1)}$ and other micro-cells $\{\mathbf{x}^{(s-1)}_m\}_m$ are computed similarly to Eqs.~\eqref{eq:6} and \eqref{eq:7}. Hence, the context-aware features of $\mathbf{x}^{(s)}$ are derived as
\begin{equation} \label{eq:12}
\mathbf{\phi}_\text{fused} = \sum_{m=1}^{|R^{(s-1)}|} f_m^{(s-1)} \, W_v^{(s-1)} \, \mathbf{\phi}(\mathbf{x}^{(s-1)}_m),
\end{equation}
where $W_v^{(s-1)}$ corresponds to learnable parameters. To preserve intrinsic features, $\mathbf{\phi}_\text{fused}$ is concatenated with the anchor cell's features, followed by a $1\times1$ convolution and ReLU activation function to reduce dimensionality and to introduce non-linearity
\begin{equation} \label{eq:13}
\mathbf{\phi}(\mathbf{x}^{(s)}) = (\mathbf{\phi}^\top_\text{fused} \quad  \mathbf{\phi}(\mathbf{x}_{a}^{(s-1)})^\top)^\top,
\end{equation}
\begin{equation} \label{eq:14}
\mathbf{\phi}_{\textrm{down}}(\mathbf{x}^{(s)}) = \mathrm{ReLU}\bigl(\mathbf{Conv}_{1\times1}( \mathbf{\phi}(\mathbf{x}^{(s)}) ) \bigr),
\end{equation}
where $\mathbf{\phi}_{\textrm{down}}(\mathbf{x}^{(s)})$ serves as the refined feature representation for the macro-cell $\mathbf{x}^{(s)}$. This procedure, defined by Eqs.~\eqref{eq:12}-\eqref{eq:14}, is repeated through all scales in $\{R^{(s)}\}_{s=1}^S$.

\subsection{Panoptic Context Aggregation Network (PanCAN)} \label{subsec:multiscaleorder}

As illustrated in Fig.~\ref{fig:framework}, the proposed network is composed of $S$ modules, each corresponding to a distinct resolution level. The output of a module operating on a finer resolution serves as the input to the module for the next coarser resolution. Within each module, multi-order context-aware features are constructed through $T$ layers, where $T$ is empirically determined as described in the experimental section. Given a cell $\mathbf{x}^{(s)}$ at scale $s$, its multi-order context-aware features from different  $c \in \{1,\dots,C\}$ are concatenated as
\begin{equation} \label{eq:16}
\mathbf{\phi}_{c}(\mathbf{x}^{(s)}) = \left(\mathbf{\phi}_{c,1}(\mathbf{x}^{(s)})^{\top} \quad \mathbf{\phi}_{c,2}(\mathbf{x}^{(s)})^{\top} \quad \ldots\right)^{\top},
\end{equation}
where each component $\mathbf{\phi}_{c,k}(\mathbf{x}^{(s)})$ is evaluated by Eq.~\eqref{eq:8}. Therefore, the comprehensive context-aware features for a given cell $\mathbf{x}^{(s)}$ are obtained by concatenating features through all values of $c \in \{1,\dots,C\}$, as per Eq.~\eqref{eq:4}:
\begin{equation} \label{eq:17}
\small
\phi(\x^{(s)}) = \bigg( {\phi^{(0)}(\x^{(s)})}^{\top}  \ \gamma^{\frac{1}{2}} \P_{1}^{(s,t)}  {\phi_{1}(\x^{(s)})}^{\top}  \  \ldots \ \gamma^{\frac{1}{2}} \P_{C}^{(s,t)} {\phi_{c}(\x^{(s)})}^{\top}\bigg),
\end{equation}
where $\{\mathbf{P}_{c}^{(s,t)}\}_{c=1}^C$ are the learned neighborhood relationships for the $t$-th layer within the $s$-th module.

It is clear that the dimensionality of $\mathbf{\phi}(\mathbf{x}^{(s)})$ increases with the number and order of context types, as well as the network's depth. To mitigate the curse of dimensionality, we employ a $1\times1$ convolution within each module to reduce dimensionality~\cite{li2021more}:
\begin{equation} \label{eq:19}
    \mathbf{\psi}(\mathbf{x}^{(s)}) = \mathbf{Conv}_{1\times1} \big( \mathbf{\phi}(\mathbf{x}^{(s)}) \big),
\end{equation}
where $\mathbf{\phi}(\mathbf{x}^{(s)})$ and $\mathbf{\psi}(\mathbf{x}^{(s)})$ denote the full context-aware features and their dimensionality-reduced counterparts, respectively.

After computing the multi-order context-aware features for all cells $\mathbf{x}^{(s)}$ at scale $s$, for each macro-cell $\mathbf{x}^{(s+1)}$ at scale $(s+1)$, its anchor micro-cell $\mathbf{x}^{(s)}_a$ from scale $s$ is estimated based on its saliency. Subsequently, attention weights between the anchor micro-cell and other micro-cells are evaluated, and the features for the macro-cell $\mathbf{x}^{(s+1)}$ are derived using Eqs.~\eqref{eq:12}-\eqref{eq:14}.

In the proposed network, hierarchical feature maps are generated through progressive aggregation from micro-cells to macro-cells. Each module generates features $\{ \mathbf{\psi}(\mathbf{x}^{(s)}) \}$. Additionally, a global feature $\mathbf{\psi}^{(0)}$ is incorporated, obtained by aggregating all original cells without initial partitioning. The features for an image $\mathcal{I}_p$ at scale $s$ are aggregated across all its cells as:
\begin{equation}\label{eq:img_con_feats}
\mathbf{\phi}_{\mathcal{K}}^{(s)}(\mathcal{I}_p) = \sum_{i=1}^{N_h} \sum_{j=1}^{N_w} \mathbf{\psi}(\mathbf{x}^{(s)}_{i,j}), \quad \forall \mathbf{x}^{(s)}_{i,j} \in R^{(s)}.
\end{equation}

To combine features across scales, a multi-head attention mechanism is employed:
\begin{equation} \label{eq:15}
\phi_{\K}^{ms}(\I_p) = \mathrm{MultiHead}\left(\phi_{\K}^{(1)}(\I_p), \  \ldots, \ \phi_{\K}^{(S)}(\I_p), \ \phi_{\K}^{(0)}(\I_p) \right),
\end{equation}
where $\mathbf{\phi}_{\mathcal{K}}^{(0)}(\mathcal{I}_p)$ represents the aggregated features of the image without initial partitioning. Finally, a fully connected layer is placed atop the network for multi-label image classification.
 \subsection{End-to-end Supervised Learning} \label{subsec:learningalgo}

As depicted in Fig.~\ref{fig:framework}, the proposed network is constructed in an end-to-end fashion, designed to learn cross-scale and multi-order context-aware features for robust multi-label classification. The network parameters comprise the neighborhood relationships $\{\mathbf{P}_{c}^{(s, t)}\}_{c=1}^C$ and the attention weights across different modules. The training procedure adopts a grouped strategy to address label co-occurrence and class imbalance, enabling the model to robustly capture complex label dependencies. Given $N$ training images $\{\mathcal{I}_p\}_{p=1}^N$ and their corresponding category labels $Y_p^l$ (where $Y_p^l=1$ if $\mathcal{I}_p$ belongs to the $l^\textrm{th}$ category, and $Y_p^l=-1$ otherwise), to mitigate class imbalance, the total loss is structured into $G$ groups of loss values, with each group comprising $N_g$ training samples based on label co-occurrence. The loss function is defined as:
\begin{align} \label{eq:context-awareloss2}
    \min_{\{W_g\}, \{\mathbf{P}_{c}^{(s,t)}\}} & \frac{1}{2} \sum_{g=1}^{G} \left( \|W_g\|_2^2 + C_g \sum_{p=1}^{N_g} \mathcal{L}_g(W_g \mathbf{\phi}_{\mathcal{K}}^{ms}(\mathcal{I}_p), Y_{p, g}^l) \right).
\end{align}
Here, $W_g$ denotes the weight matrix for the fully connected layer, $\{\mathbf{P}_c^{s,t}\}_c$ represents the learnable neighborhood relationships, and $C_g$ is a hyper-parameter. $\mathcal{L}_g$ is the cross-entropy function, and $\ell_2$-norm regularization is applied to the weights $\{W_g\}$ across the groups. The network is optimized using error backpropagation and a gradient descent algorithm.

\section{Experiments} \label{sec:experiments}

In this section, we evaluate the proposed networks on three benchmark datasets (i.e.~NUS-WIDE~\cite{chua2009nus}, PASCAL VOC2007~\cite{everingham2010pascal}, and MS-COCO~\cite{lin2014microsoft}). Unless otherwise specified, for the training procedure in all the experiments, 200 epochs of learning iterations are set by using the AdamW optimizer with a batch size of 6 and a maximum learning rate of $0.0001$. An early stopping strategy and exponential moving average with a decay rate of $0.9997$ are also employed, along with data augmentation techniques such as RandAugment~\cite{cubuk2020randaugment} and Cutout~\cite{devries2017improved}.

\subsection{Results on the NUS-WIDE Benchmark} \label{sec:experiments:nuswide}
The NUS-WIDE dataset is a widely used benchmark for multi-label image classification, comprising 269,648 Flickr images with 5,018 labels, manually annotated with 81 specific concepts, averaging 2.4 concepts per image. According to the official division, 161,789 images are used for training and 107,859 for testing, with small-size images selected for our experiments. All images were resized to 400×500 pixels, and the context structure was partitioned based on an initial 8×10 grid. For feature extraction, we respectively applied the pre-trained models including ResNet101~\cite{he2016deep}, Tresnet~\cite{ridnik2021tresnet}, and Cvt~\cite{wu2021cvt} trained on the ImageNet dataset for the input of the proposed context-aware network, showing the robustness to the feature initializations. The performance are measured in three metrics: mean Average Precision (mAP), Combined F1 score (CF1), and Overall F1 score (OF1), where higher scores indicate better performance.

\subsubsection{Ablation study} 

To optimize the network architecture, we conducted a comprehensive ablation study of each module on the NUS-WIDE dataset. Firstly, we evaluated the performance impact of individual components: the multi-order module, the cross-scale module, the grouped fully connected layer, and the random walk within the multi-order module. Secondly, we performed a detailed analysis of various strategies, specifically examining the context-aware distance, sub-region selection methods, and the random walk's probability threshold. All experiments in this study utilized a pre-trained ResNet101 as the feature extractor, processing images on an 8$\times$10 grid for input.
\begin{table}[t]
	\centering
	\caption{Ablation study on context-aware module with different context orders}
	\label{tab:context_aware_ablation}
	\begin{tabular}{cccc}
		\toprule
		Neighborhood Order & mAP & CF1 & OF1 \\ 
		\midrule
		First-Order  & 66.5 & 64.9 & 75.0 \\
		Second-Order & \textbf{66.9} & \textbf{65.2} & \textbf{75.5}  \\
		Third-Order  & 66.8 & 65.0 & 75.3 \\
		\bottomrule
	\end{tabular}
\end{table}

\begin{table}[t]
	\centering
	\caption{Ablation study on the number of context-aware layer depth in a scale}
	\label{tab:multi_scale_ablation}
	\begin{tabular}{cccc}
		\toprule
		Layer Number  & mAP & CF1 & OF1 \\ 
		\midrule
		One-Layer     & 66.4 & 64.6 & 75.1 \\
		Two-Layer    & 66.7 & 65.1 & 75.3 \\
		Three-Layer  & \textbf{66.9} & \textbf{65.2} & \textbf{75.5}  \\
		\bottomrule
	\end{tabular}
\end{table}

\begin{table}[t]
	\centering
	\caption{Ablation study on random walk with different threshold values, \ding{55} means that all the cells are considered.}
	\label{tab:random_walk_ablation}
	\begin{tabular}{cccc}
		\toprule
		Threshold Value & mAP & CF1 & OF1 \\
		\midrule
		\ding{55} & 66.4 & 64.8 & 75.3 \\
		0.62      & 66.7 & 65.0 & 75.4 \\
		0.67      & 66.8 & 65.0 & 75.6 \\
		0.71      & \textbf{66.9} & \textbf{65.2} & \textbf{75.5}  \\
		0.75      & 66.7 & 65.1 & 75.3 \\
		\bottomrule
	\end{tabular}
\end{table}

\begin{table}[t]
	\centering
	\caption{Ablation study on scale interval between two consecutive scales}
	\label{tab:subregion_size_ablation}
	\begin{tabular}{cccc}
		\toprule
		Scale interval & mAP & CF1 & OF1 \\
		\midrule
		$1\times1$  & 66.4 & 64.7 & 74.9 \\
		$2\times2$  & \textbf{66.9} & \textbf{65.2} & \textbf{75.5} \\
		$3\times3$  & 66.7 & 64.9 & 75.3 \\
		\bottomrule
	\end{tabular}
\end{table}

\begin{table}[t]
	\centering
	\caption{Performance impact of each module for grid of $8\times10$ cells on NUS-WIDE dataset}
	\label{tab:module_ablation}
	\begin{tabular}{cccc}
		\toprule
		Configuration & mAP & CF1 & OF1 \\
		\midrule
		PanCAN & \textbf{66.9} & \textbf{65.2} & \textbf{75.5} \\
		PanCANk without Multi-order  & 66.1 & 64.6 & 74.7 \\
		PanCAN without Cross-scale  & 66.3 & 64.9 & 74.9 \\
		PanCAN without Grouped FC   & 66.7 & 65.0 & 75.2 \\
		PanCAN without Random walk  & 66.6 & 64.8 & 75.3 \\
		\bottomrule
	\end{tabular}
\end{table}

\begin{table}[t]
    \centering
    \caption{Performance comparison of different methods on the NUS-WIDE dataset. For the methods with the same backbone network, the best results are shown in red and the second best results are in blue. ``PanCAN'' stands for the proposed network.} \label{tab:nuswideperformance}
    \resizebox{0.98\linewidth}{!}{
	\begin{tabular}{lccccc}
		\toprule
		\textbf{Method} & \textbf{Backbone Network} & \textbf{Cells} & \textbf{mAP} & \textbf{CF1} & \textbf{OF1}\\
		\midrule
		MS-CMA \cite{you2020cross}     & ResNet101   & -        & 61.4 & 60.5 & 73.8 \\
		SRN \cite{zhu2017learning}     & ResNet101   & -        & 62.0 & 58.5 & 73.4 \\
		ICME \cite{chen2019multi}      & ResNet101   & -        & 62.8 & 60.7 & 74.1 \\
		ASL \cite{ridnik2021asymmetric}  & ResNet101   & -      & 65.2 & 63.6 & 75.0 \\
		Q2L-R101 \cite{liu2021query2label} & ResNet101 & -      & 65.0 & 63.1 & 75.0 \\
		ML-SGM \cite{wu2023semantic}   & ResNet101   & -        & 64.6 & 62.4 & 72.5 \\
		SST \cite{chen2022sst}         & ResNet101   & -        & 63.5 & 59.6 & 73.2 \\
		SADCL \cite{ma2023semantic}    & ResNet101   & -        & 65.9 & 63.0 & \textcolor{blue}{75.0} \\
		\multirow{2}{*}{MCDKN~\cite{JiuZhuicpr2024}}         & ResNet101   & 4$\times$5  & 65.4 & 63.9 & 74.2 \\
		& ResNet101   & 8$\times$10 & \textcolor{blue}{66.3} & \textcolor{blue}{64.6} & 74.8 \\
		\multirow{2}{*}{PanCAN} & ResNet101   & 4$\times$5 & 66.1 & 64.5 & 74.9 \\
		& ResNet101   & 8$\times$10 & \textcolor{red}{66.9} & \textcolor{red}{65.2} & \textcolor{red}{75.5} \\
		\midrule
		Focal loss \cite{lin2017focal}   & TResNetL    & -        & 64.0 & 62.9 & 74.7 \\
		ASL \cite{ridnik2021asymmetric}  & TResNetL    & -        & 65.2 & 63.6 & 75.0 \\
		Q2L-TResL \cite{liu2021query2label}   & TResNetL    & -        & 66.3 & 64.0 & 75.0 \\
		\multirow{2}{*}{MCDKN~\cite{JiuZhuicpr2024}}  & TResNetL    & 4$\times$5  & 66.9 & 64.5 & 75.8 \\
		& TResNetL    & 8$\times$10 & \textcolor{blue}{67.8} & \textcolor{blue}{65.1} & \textcolor{blue}{76.5} \\
		\multirow{2}{*}{PanCAN}  & TResNetL    & 4$\times$5 & 67.6 & 65.1 & 76.9 \\
		& TResNetL    & 8$\times$10 & \textcolor{red}{68.3} & \textcolor{red}{65.7} & \textcolor{red}{77.6} \\
		\midrule
		MlTr-l\cite{cheng2022mltr}    & MlTr-l(22k) & -        & 66.3 & 65.0 & 75.8\\
		Q2L-CvT  \cite{liu2021query2label}  & CvT-W24     & -    & \textcolor{blue}{70.1} & 67.6 & 76.3 \\
		\multirow{2}{*}{MCDKN~\cite{JiuZhuicpr2024}}   & CvT-W24    & 4$\times$5  & 69.4 & 68.2 & 76.1 \\
		&CvT-W24     & 8$\times$10 & 69.7 & \textcolor{blue}{68.9} & \textcolor{blue}{76.6} \\
		\multirow{2}{*}{PanCAN}   & CvT-W24    & 4$\times$5 & 69.8 & 69.6 & 77.5 \\
		& CvT-W24    & 8$\times$10 & \textcolor{red}{70.4} & \textcolor{red}{69.9} & \textcolor{red}{77.8} \\
		\bottomrule
	\end{tabular}
	}
\end{table}

\begin{table*}[t]
	\centering
	\caption{Performance comparison of different methods on the PASCAL VOC2007 dataset. ``PanCAN'' stands for the proposed network. The best results are shown in bold. }
	\label{tab-PASCAL VOC}
	\resizebox{\linewidth}{!}{
	\begin{tabular}{lccccccccccccccccccccc}
	\hline
	\textbf{Method} & \textbf{aero} & \textbf{bike} & \textbf{bird} & \textbf{boat} & \textbf{bottle} & \textbf{bus} & \textbf{car} & \textbf{cat} & \textbf{chair} & \textbf{cow} & \textbf{table} & \textbf{dog} & \textbf{horse} & \textbf{mbike} & \textbf{person} & \textbf{plant} & \textbf{sheep} & \textbf{sofa} & \textbf{train} & \textbf{tv} & \textbf{mAP} \\
	\toprule
	CNN-RNN \cite{wang2016cnn} & 96.7 & 83.1 & 94.2 & 92.8 & 61.2 & 82.1 & 89.1 & 94.2 & 64.2 & 83.6 & 70.0 & 92.4 & 91.7 & 84.2 & 93.7 & 59.8 & 93.2 & 75.3 & 99.7 & 78.6 & 84.0 \\
        
	VGG+SVM \cite{simonyan2014very} & 98.9 & 95.0 & 96.8 & 95.4 & 69.7 & 90.4 & 93.5 & 96.0 & 74.2 & 86.6 & 87.8 & 96.0 & 96.3 & 93.1 & 97.2 & 70.0 & 92.1 & 80.3 & 98.1 & 87.0 & 89.7 \\ 
        
	Fev+Lv \cite{yang2016exploit} & 97.9 & 97.0 & 96.6 & 94.6 & 73.6 & 93.9 & 96.5 & 95.5 & 73.7 & 90.3 & 82.8 & 95.4 & 97.7 & 95.9 & 98.6 & 77.6 & 88.7 & 78.0 & 98.3 & 89.0 & 90.6 \\ 
        
	HCP \cite{wei2015hcp} & 98.6 & 97.1 & 98.0 & 95.6 & 75.3 & 94.7 & 95.8 & 97.3 & 73.1 & 90.2 & 80.0 & 97.3 & 96.1 & 94.9 & 96.3 & 78.3 & 94.7 & 76.2 & 97.9 & 91.5 & 90.9 \\ 
        
	RDAL \cite{wang2017multi} & 98.6 & 97.4 & 96.3 & 96.2 & 75.2 & 92.4 & 96.5 & 97.1 & 76.5 & 92.0 & 87.7 & 96.8 & 97.5 & 93.8 & 98.5 & 81.6 & 93.7 & 82.8 & 98.6 & 89.3 & 91.9 \\ 
        
	RARL \cite{chen2018recurrent} & 98.6 & 97.1 & 97.1 & 95.5 & 75.6 & 92.8 & 96.8 & 97.3 & 78.3 & 92.2 & 87.6 & 96.9 & 96.5 & 93.6 & 98.5 & 81.6 & 93.1 & 83.2 & 98.5 & 89.3 & 92.0 \\ 
        
	SSGRL \cite{chen2019learning} & 99.5 & 97.1 & 97.6 & 97.8 & 82.6 & 94.8 & 96.7 & 98.1 & 78.0 & 97.0 & 85.6 & 97.8 & 98.3 & 96.4 & 98.1 & 84.9 & 96.5 & 79.8 & 98.4 & 92.8 & 93.4 \\
        
	MCAR \cite{gao2021learning} & 99.7 & 99.0 & 98.5 & 98.2 & 85.4 & 96.9 & 97.4 & 98.9 & 83.7 & 95.5 & 88.8 & 99.1 & 98.2 & 95.1 & 99.1 & 84.8 & 97.1 & 87.8 & 98.3 & 94.8 & 94.8 \\ 
        
	ASL \cite{ridnik2021asymmetric} & 99.9 & 98.4 & 98.9 & 98.7 & 86.8 & 98.2 & 98.7 & 98.5 & 83.1 & 98.3 & \textbf{89.5} & 98.8 & 99.2 & 98.6 & 99.3 & 89.5 & 99.4 & 86.8 & \textbf{99.6} & 95.2 & 95.8 \\ 
        
	ADD-GCN \cite{ye2020attention} & 99.8 & 99.0 & 98.4 & \textbf{99.0} & 86.7 & 98.1 & 98.5 & 98.3 & \textbf{85.8} & 98.3 & 88.9 & 98.8 & 99.0 & 97.4 & 99.2 & 88.3 & 98.7 & \textbf{90.7} & 99.5 & \textbf{97.0} & 96.0 \\ 
        
	Q2L-TResL \cite{liu2021query2label} & \textbf{99.9} & 98.9 & 99.0 & 98.4 & \textbf{87.7} & 98.6 & 98.8 & 99.1 & 84.5 & 98.3 & 89.2 & 99.2 & 99.2 & 99.2 & 99.3 & \textbf{90.2} & 98.8 & 88.3 & 99.5 & 95.5 & 96.1 \\ 
        
	PanCAN  & 99.8 & \textbf{99.5} & \textbf{99.3} & 98.9 & 87.5 & \textbf{98.8} & \textbf{99.2} & \textbf{99.4} & 84.7 & \textbf{98.6} & 89.2 & \textbf{99.3} & \textbf{99.4} & \textbf{99.6} & \textbf{99.8} & 89.9 & \textbf{99.5} & 89.3 & 99.5 & 96.7 & \textbf{96.4} \\
	\bottomrule
    \end{tabular}
	}
\end{table*}

\begin{table*}[t]
	\centering
	\caption{Performance comparison of different methods on the MS-COCO dataset. ``PanCAN'' stands for the proposed network. The best results for each type of backbone network are shown in bold.}
	\label{MS-COCO}
	\resizebox{\textwidth}{!}{
		\begin{tabular}{lcccccccccccccccccccccc}
			\hline
			\multirow{2}{*}{\textbf{Method}} & \textbf{Backbone} & \multirow{2}{*}{\textbf{Resolution}} & 
			\multirow{2}{*}{\textbf{mAP}} & \multirow{2}{*}{\textbf{CP}} & \multirow{2}{*}{\textbf{CR}} & \multirow{2}{*}{\textbf{CF1}} &
			\multirow{2}{*}{\textbf{OP}} & \multirow{2}{*}{\textbf{OR}} & \multirow{2}{*}{\textbf{OF1}} &
			\textbf{CP} & \textbf{CR} & \textbf{CF1} &
			\textbf{OP} & \textbf{OR} & \textbf{OF1} \\ 
			 & \textbf{Network}  &  &  &  &  &  & &  &  &
			\textbf{(Top 3)} & \textbf{(Top 3)} & \textbf{(Top 3)} &
			\textbf{ (Top 3)} & \textbf{(Top 3)} & \textbf{(Top 3)} \\ 
			\toprule
			SRN \cite{zhu2020deformable} & ResNet101 & 224$\times$224 & 
			77.1 & 81.6 & 65.4 & 71.2 & 82.7 & 69.9 & 75.8 &
			85.2 & 58.8 & 67.4 & 87.4 & 62.5 & 72.9 \\ 
			
			ResNet-101 \cite{he2016deep} & ResNet101 & 224$\times$224 &
			78.3 & 80.2 & 66.7 & 72.8 & 83.9 & 70.8 & 76.6 &
			84.1 & 59.4 & 69.7 & 89.1 & 62.8 & 73.6 \\ 
			
			CADM \cite{chen2019multi} & ResNet101 & 448$\times$448 &
			82.3 & 85.2 & 72.0 & 78.0 & 85.6 & 74.6 & 79.6 &
			87.1 & 63.6 & 73.5 & 89.4 & 66.0 & 76.0 \\ 
            
			ML-GCN \cite{chen2019multi} & ResNet101 & 448$\times$448 &
			83.0 & 85.1 & 72.2 & 78.0 & 85.1 & 74.7 & 79.4 &
			86.3 & 63.4 & 74.2 & 89.1 & 66.7 & 76.3 \\ 
            
			KSSNet \cite{liu2018multi} & ResNet101 & 448$\times$448 &
			83.7 & 84.1 & 73.1 & 78.2 & \textbf{87.8} & 76.2 & 81.5 &
			- & - & - & - & - & - \\ 
            
			MS-CMA \cite{you2020cross} & ResNet101 & 448$\times$448 &
			83.8 & 84.9 & 74.4 & 78.4 & 84.4 & 77.9 & 81.0 &
			86.7 & 64.9 & 74.3 & 90.9 & 67.2 & 77.2 \\ 
            
			MCAR \cite{gao2020multi} & ResNet101 & 448$\times$448 &
			83.8 & 84.1 & 74.8 & 79.3 & 87.1 & 78.0 & 83.1 &
			88.1 & 65.7 & 75.9 & 90.6 & 66.3 & 76.7 \\ 
            
			SSGRL \cite{chen2019learning}& ResNet101 & 576$\times$576 &
			83.8 & 85.9 & 74.8 & 80.0 & 87.7 & 79.1 & \textbf{83.7} &
			\textbf{91.9} & 62.5 & 72.7 & \textbf{93.8} & 64.1 & 76.2 \\
            
			C-Trans \cite{lanchantin2021general} & ResNet101 & 576$\times$576 &
			85.1 & 84.3 & 74.9 & 79.7 & 87.7 & 76.5 & 81.7 &
			90.1 & 65.7 & 76.0 & 92.1 & \textbf{71.4} & 77.6 \\
            
			ADD-GCN \cite{ye2020attention} & ResNet101 & 576$\times$576 &
			85.7 & 84.8 & 74.5 & 79.6 & 86.6 & 76.9 & 82.0 &
			86.6 & 65.6 & 75.8 & 90.6 & 70.5 & 77.9 \\ 
            
			\multirow{2}{*}{Q2L-R101 \cite{liu2021query2label}} & ResNet101 & 448$\times$448 &
			84.9 & 84.8 & 74.5 & 79.3 & 86.6 & 76.9 & 81.5 &
			78.0 & \textbf{69.1} & 73.3 & 80.7 & 70.8 & 75.4 \\ 
			& ResNet101 & 576$\times$576 &
			86.5 & 85.8 & 76.7 & 81.0 & 87.0 & \textbf{79.8} & 82.8 &
			90.4 & 66.3 & 76.5 & 92.4 & 67.9 & 78.3 \\ 
            
			PanCAN & ResNet101 & 576$\times$576 &
			\textbf{86.9} & \textbf{86.1} & \textbf{77.2} & \textbf{81.4} & 87.2 & 79.5 & 82.9 &
			90.6 & 66.5 & \textbf{76.7} & 92.6 & 68.5 & \textbf{78.5} \\ 
			\midrule
			
			ASL \cite{ridnik2021asymmetric} & TResNet-L & 448$\times$448 &
			86.6 & 87.2 & 76.4 & 81.4 & 88.2 & 79.2 & 81.8 &
			91.8 & 63.4 & 75.1 & 92.9 & 66.4 & 77.4 \\ 
			
			\multirow{2}{*}{Q2L-TResL \cite{liu2021query2label}} & TResNet-L & 448$\times$448 &
			87.3 & \textbf{87.6} & 76.5 & 81.6 & \textbf{88.4} & 78.5 & 83.1 &
			91.9 & 66.2 & 77.0 & \textbf{93.5} & 67.6 & 78.5 \\ 
			
			& TResNet-L (22k) & 448$\times$448 &
			89.2 & 86.3 & 81.4 & 83.8 & 86.5 & 83.3 & \textbf{84.9} &
			91.6 & 69.4 & 79.0 & 92.9 & 70.5 & \textbf{80.2} \\ 
            
			PanCAN & TResNet-L & 448$\times$448 &
			\textbf{90.1} & 87.2 & \textbf{82.3} & \textbf{84.6} & 87.2 & \textbf{84.9} & 84.7 &
			\textbf{92.5} & \textbf{69.8} & \textbf{79.5} & \textbf{93.5} & \textbf{71.0} & 80.0 \\ 
			\midrule
            
			MlTr-l \cite{cheng2022mltr} & MlTr-l (22k) & 384$\times$384 &
			88.5 & 86.0 & 81.7 & 83.4 & 84.9 & 82.1 & 83.9 &
			93.6 & 69.9 & 80.0 & 94.3 & 71.1 & 81.1 \\ 
            
			Swin-L \cite{liu2021swin} & Swin-L (22k) & 384$\times$384 &
			89.5 & 86.9 & \textbf{89.4} & 81.6 & 86.3 & 83.1 & 84.7 &
			93.9 & 70.8 & 81.3 & 94.1 & 71.5 & 81.3 \\ 
            
			CvT-W24 \cite{wu2021cvt} & CvT-W24 (22k) & 384$\times$384 &
			90.5 & \textbf{89.4} & 81.7 & 84.5 & 89.2 & 84.6 & 86.8 &
			\textbf{94.9} & 71.4 & 81.7 & 94.1 & 71.7 & 81.3 \\ 
			\multirow{2}{*}{Q2L-SwinL \cite{liu2021query2label}} & Swin-L (22k) & 384$\times$384 &
			90.5 & \textbf{89.4} & 81.7 & 84.5 & 89.2 & 84.6 & 86.8 &
			93.9 & 70.5 & 80.5 & 94.8 & 71.0 & 81.2 \\ 
			& CvT-W24(22k)  & 384$\times$384 & 91.3 & 88.8 & 83.2 & 85.9 & 89.2 & 84.6 & 86.8 & 93.9 & 72.1 & 81.6 & 94.7 & 72.1 & 81.6 \\ 
            
			PanCAN & CvT-W24 (22k) & 384$\times$384 &
			\textbf{92.2} & \textbf{89.4} & 83.9 & \textbf{86.6} & \textbf{89.7} & \textbf{85.0} & \textbf{87.5} & 94.8 & \textbf{72.6} & \textbf{82.3} & \textbf{95.4} & \textbf{72.5} & \textbf{82.3} \\
			\bottomrule
		\end{tabular}
	}
\end{table*}

\noindent \textbf{Impact of orders:} Here we study the impact of different neighborhood orders. Tab.~\ref{tab:context_aware_ablation} shows the performance of the context-aware module by using different neighborhood orders. It is observed that the second-order neighborhood achieves the best results, improving mAP by 0.4, CF1 by 0.3, and OF1 by 0.5 compared to the first-order neighborhood. However, it is interesting to note that the performance drops slightly for the third-order neighborhood, the reason might come from the fact that for the centering cell, the third-order neighborhood covers almost half of the image for $8\times10$ cells, causing redundancy noise \textcolor{black}{(clutter)} and then hindering the performance. In the following experiments, two-order neighborhood is adopted for the multi-order context-aware module.

\noindent \textbf{Impact of context depth:} Table~\ref{tab:multi_scale_ablation} shows the impact of \textcolor{black}{context layer} depth (T) within a single scale, utilizing second-order neighborhoods. Performance consistently improved from one to three layers: mAP from 66.4 to 66.9, CF1 from 64.6 to 65.2, and OF1 from 75.1 to 75.5. This indicates that a deeper context network effectively integrates both global and local features. However, the marginal nature of these gains suggests potential challenges, such as vanishing gradients or over-smoothing, with further increases in depth.

\noindent \textbf{Impact of random walk selection:} We investigate cell connections via random walk within the multi-order neighborhood. Table~\ref{tab:random_walk_ablation} details the effect of various random walk thresholds. Performance steadily improved from no constraint (mAP 66.4) to a peak at a threshold of 0.71 (mAP 66.9, CF1 65.2, OF1 75.5). This highlights the importance of moderate thresholding for enhancing feature interactions. Conversely, a higher threshold of 0.75 slightly degraded performance, indicating that excessive pruning may prevent useful information propagation. Based on these findings, an empirical threshold of 0.71 is adopted for the random walk in subsequent experiments. \\
\noindent \textbf{Impact of scale interval:} We study the impact of different scale interval sizes, where a macro-cell at the $s$-th scale is composed of a grid of micro-cells at the $(s-1)$-th scale. Table~\ref{tab:subregion_size_ablation} presents the results for various scale intervals. A 2$\times$2 interval achieved the highest performance across all metrics (mAP 66.9, CF1 65.2, OF1 75.5), demonstrating a balance between local structure and global context integration. In contrast, a 1$\times$1 interval performed the worst due to its lack of local detail. A 3$\times$3 interval yielded slightly inferior performance compared to 2$\times$2, likely due to the presence of additional background noise. These results highlight the importance of appropriate granularity for the scale interval.\\
\noindent \textbf{Impact of each module:} We investigate the impact of different modules in the proposed network. Tab.~\ref{tab:module_ablation} shows the results of different modules on the NUS-WIDE dataset on the grid of $8\times10$ cells. The full network achieves the best performance across all metrics (mAP 66.9, CF1 65.2, OF1 75.5), and then we remove one module and re-train the network. It can be seen that: firstly, the network without multi-order context-aware module results in a noticeable drop in mAP and OF1 by 0.8, and CF1 by 0.6, demonstrating its crucial role in capturing large contextual dependencies. Secondly, the network without cross-scale context-aware module causes a moderate decline (mAP 66.3, CF1 64.9, OF1 74.9), indicating that cross-scale aggregation is beneficial. Thirdly, the grouped FC module's removal leads to marginal decrease in CF1 from 65.2 to 65.0, reflecting its utility in enhancing category-level prediction via label grouping. Lastly, random walk module has the smallest impact, with mAP dropping to 66.6 and OF1 to 75.3, implying its complementary role in refining feature relationships.

\subsubsection{Performance comparison}

\textcolor{black}{According to the preceding ablation study, we adopt two initial grid sizes for each image: $4\times5$ and $8\times10$. The $4\times5$ configuration generates a cross-scale hierarchy including $4\times5$, $2\times3$, $1\times2$, and $1\times1$ cell grids. The $8\times10$ configuration extends the aforementioned hierarchy by introducing an additional finer scale level, thereby enabling the modeling of richer low-level structures. For each scale, we employ a second-order neighborhood and three layer networks to obtain contextual representation. Notably, for the coarse scales (specifically $2\times3$ and $1\times2$), we simplify the model by applying only a first-order neighborhood to reduce complexity and prevent redundant modeling.}  

Tab.~\ref{tab:nuswideperformance} shows the comparison performance of the proposed network with other state-of-the-art methods on the NUS-WIDE dataset. The best performance is shown in red and the second in blue. It can be observed that: i) when using ResNet101 as the backbone network, PanCAN achieves 66.9\% mAP, outperforming SADCL (65.9\% mAP) under the same configuration. Notably, CF1 improves by 2.2\%, indicating better consistency in multi-label predictions, and the OF1 index also shows a slight increase; ii) with a more powerful backbone \---TResNet-L, PanCAN further improves its performance, surpassing mainstream methods in mAP, CF1, and OF1. Specifically, mAP and CF1 increase by about 2.0\% and 1.7\%, respectively, and OF1 also improves by 2.6\% compared to Q2L-TResL; iii) when recent advanced CvT-W24 backbone is adopted, PanCAN still maintains a leading edge, outperforming the current best-performing methods across all metrics. Notably, CF1 improves by more than 2\% compared to Q2L-CvT. Therefore, it is concluded that PanCAN is able to consistently boost multi-label image classification performance across various backbone networks, which validated the robustness of the proposed network to different features. Moreover, when the grid of cells in the image increases from 4$\times$5 to 8$\times$10, the proposed networks obtain consistent gains from 4$\times$5 grid to 8$\times$10 grid under different backbones, demonstrating a strong robustness to integrate the contexts to different features.

\subsection{Results on the PASCAL VOC2007 Benchmark} \label{sec:experiments:voc}

The PASCAL Visual Object Classes (VOC) 2007 dataset is a widely recognized benchmark for object detection and classification, comprising 9,963 images annotated with 20 object categories, such as people, cars, and dogs. Each image can contain one or more labels. To ensure a fair comparison with other methods, we follow the common setting of training the model on the train-validation set and then evaluating its performance on the test set. All images are resized to $400\times500$ pixels, and an initial grid structure of $8 \times 10$ cells is employed. For feature extraction, we respectively utilize pre-trained TResNet-L model as the backbone. The proposed network with the same architecture is validated on the NUS-WIDE dataset in Section~\ref{sec:experiments:nuswide}. \\
Table~\ref{tab-PASCAL VOC} presents the performance of PanCAN compared to several state-of-the-art methods on the PASCAL VOC2007 dataset. Utilizing TResNet-L as its backbone, PanCAN achieves state-of-the-art results across most categories. Specifically, our model shows performance gains of approximately 0.2\% to 0.5\% over leading models like Q2L-TResL and ADD-GCN in key categories such as "aeroplane," "car," and "person." This significant improvement underscores PanCAN's enhanced ability to effectively capture features and variations of well-defined objects. Furthermore, PanCAN demonstrates strong robustness in complex scenes with multiple objects or occlusions; in challenging categories like "dog" and "train," our model still outperforms the best existing methods by roughly 0.1\% to 0.3\%. Overall, PanCAN attained a mean Average Precision (mAP) of 96.4\%, representing a 0.3\% improvement over the current best publicly available method, Q2L-TResL (96.1\%).
The performance improvement are mainly attributed to multi-order and cross-scale context-aware features, effectively boosting the model's ability to discriminate objects in multi-label learning tasks. Although the model shows slightly lower performance in some categories (e.g., ``plant'' and ``sofa'') due to high intra-class variance or cluttered backgrounds, where it lags behind ADD-GCN by up to 1.4\%, PanCAN still maintains strong competitive performance in these challenging categories. In conclusion, The proposed PanCAN demonstrates excellent generalization ability and robustness in the PASCAL VOC2007 multi-label recognition task, achieving state-of-the-art performance across multiple categories. In addition to quantitative results, we also present qualitative ones on PASCAL VOC2007 in Fig.~\ref{fig:Annotation_Results}. 

\begin{figure*}[tbp]
    \centering 
    \includegraphics[width=\textwidth]{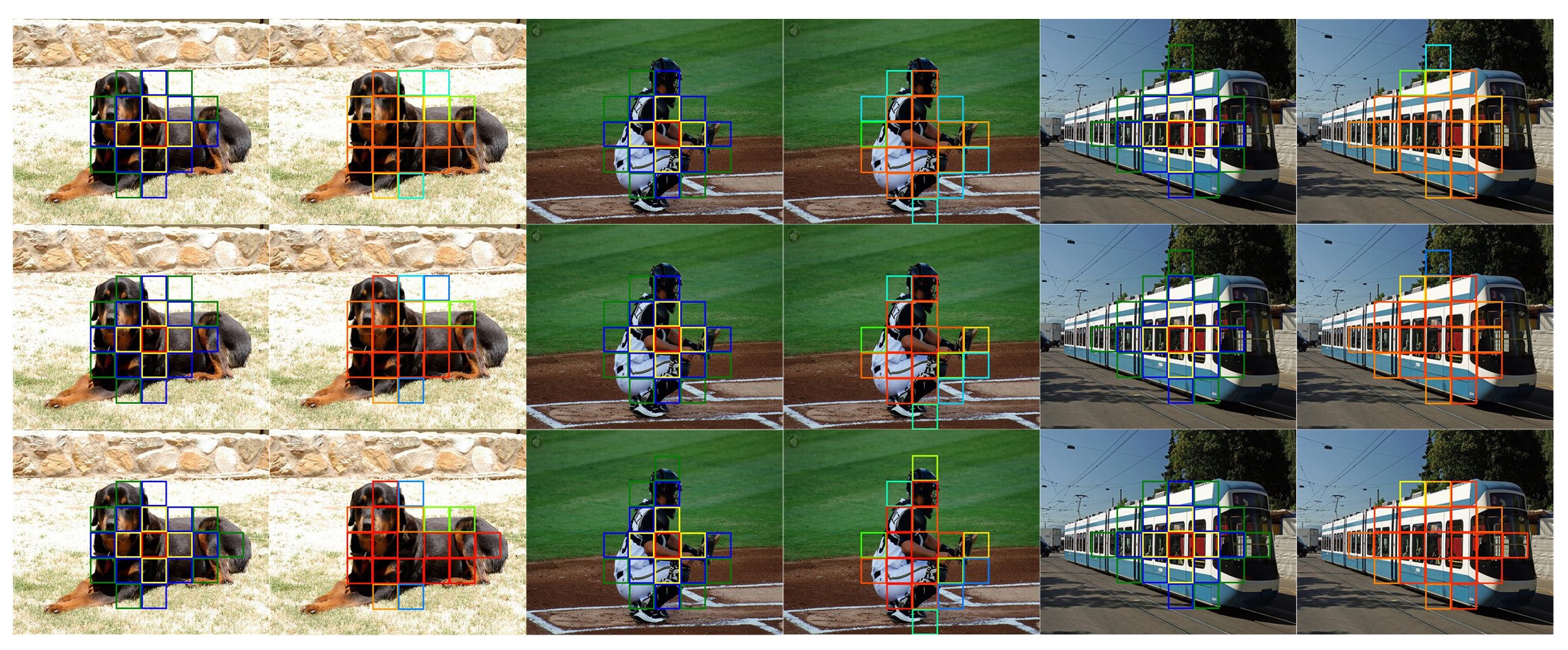}  
    \caption{\textcolor{black}{Visualization of learned contexts at the different depths of the network on NUS-WIDE dataset. The images are divided into a grid of $8\times10$ cells. From top to bottom each row respectively corresponds to the context-aware network with a depth of one, two and three layers in one scale. The 1st, 3rd and 5th columns show the learned contexts (the centering cell is marked in red, and its learned 1st-, 2nd-, and 3rd-order domains respectively are in yellow, blue and green), while the 2nd, 4th, and 6th columns visualize the learned weights of neighboring cells on the centering cell, where warmer color indicating higher influence.}} \label{fig:Visualization_of_Numbers_of_Layers}
\end{figure*}

\subsection{Results on the MS-COCO Benchmark} \label{sec:experiments:coco}

The MS-COCO dataset is primarily used for object recognition tasks in the context of scene understanding. The training set consists of 82,783 images that contain common objects in scenes. These objects are categorized into 80 classes, with an average of about 2.9 object labels per image. Since the ground-truth labels for the test set are not available, we evaluated all methods on the validation set (40,504 images). The number of labels per image varies significantly in the MS-COCO dataset. Following the commonly used evaluation metric, we filtered out labels with low probabilities when evaluating the top-3 label predictions.

In experiments, we respectively adopted three different backbone networks to extract the features for the MS-COCO dataset. Similarly, the proposed networks with the same architecture are validated on the NUS-WIDE dataset in Section~\ref{sec:experiments:nuswide}.  Tab.~\ref{MS-COCO} shows the performance comparison of our model w.r.t. various state-of-the-art methods on the MS-COCO dataset in term of mAP, CP, CR, CF1, OP, OR, and OF1 for both all-labels and top-3 predictions. From the table, it is clearly seen that PanCAN consistently achieves superior performance compared to other state-of-the-art methods. Specifically, under the ResNet-101 backbone, PanCAN demonstrated significant improvements with multiple metrics, including mAP, CF1, and OF1, surpassing methods like ADD-GCN and C-Trans. For the TResNet-L backbone, PanCAN further increases its performance, excelling particularly in category-level F1 scores and Top-3 predictions. These improvements highlight the backbone’s ability to support the model's robust adaptability to cross-scale objects in complex scenes, demonstrating how the cross-scale context fusion strategy enhances feature expressiveness and model robustness. Finally, when using the CvT-W24 backbone, PanCAN obtains the best performance across all evaluation metrics, particularly in overall precision and Top-3 label prediction, outperforming the other methods. This success can be attributed to the integration of \textcolor{black}{cross-scale and multi-order contexts}, which strengthens the consistency between fine-grained details and global semantics. In a word, these results validate that the proposed network is highly adaptable across different backbones from ResNet-101, TResNet-L, to CvT-W24, demonstrating its strong generalization and robustness for the different features; these results indeed confirm that the cross-scale and multi-order context-aware features are crucial for powerful discrimination ability in the  complex scenarios.

\begin{figure*}[tbp]
    \centering 
    \includegraphics[width=\textwidth]{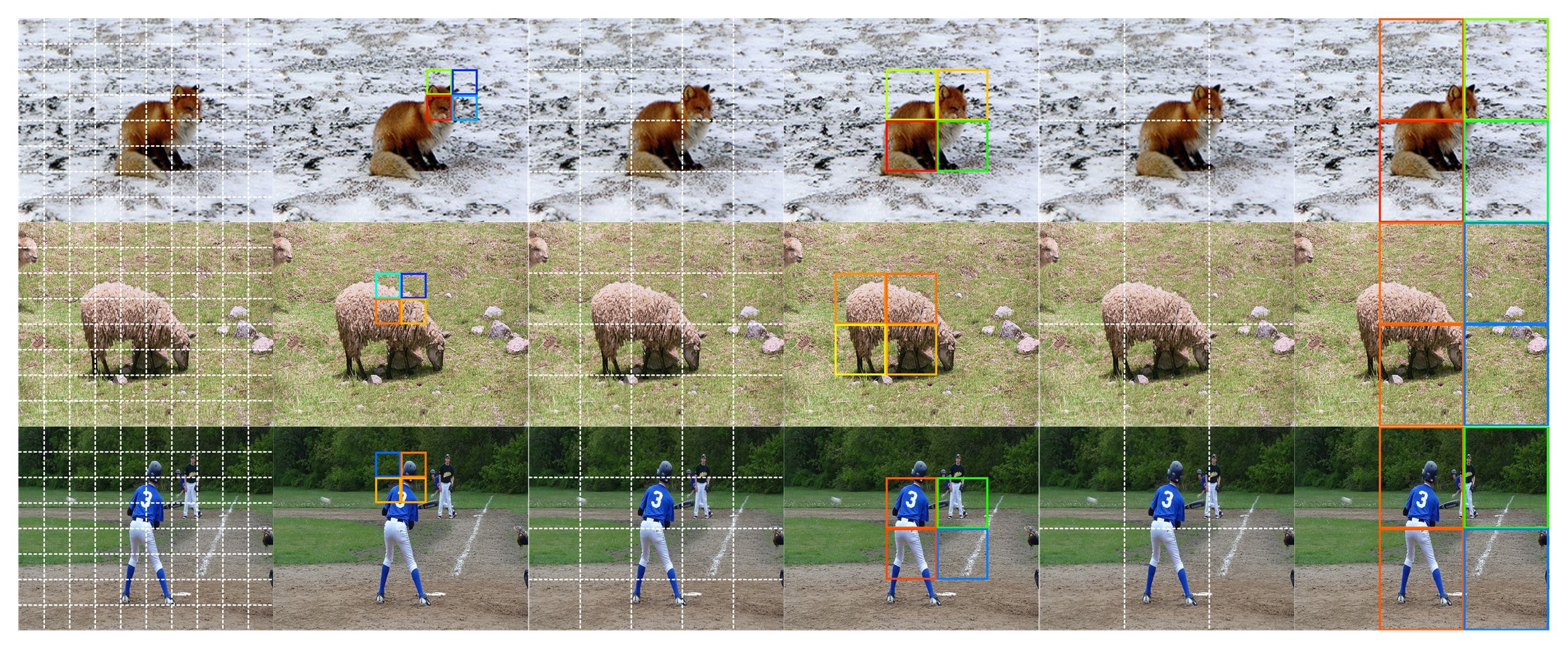}
    \caption{Visualization of learned cross-scale contexts. From top to bottom each row respectively shows one instance in the NUS-WIDE, PASCAL VOC2007, MS-COCO dataset. The 1st, 3rd an 5th columns show the grids at different scales separated by white dashed lines. The 2nd, 4th and 6th columns illustrate the influence of the micro-cell on the features of macro-cell learned by the network where warmer color stands for higher importance.} \label{fig:Visualization_of_learning_results}
\end{figure*}

\begin{figure*}[tbp]
    \centering 
    \includegraphics[width=\textwidth]{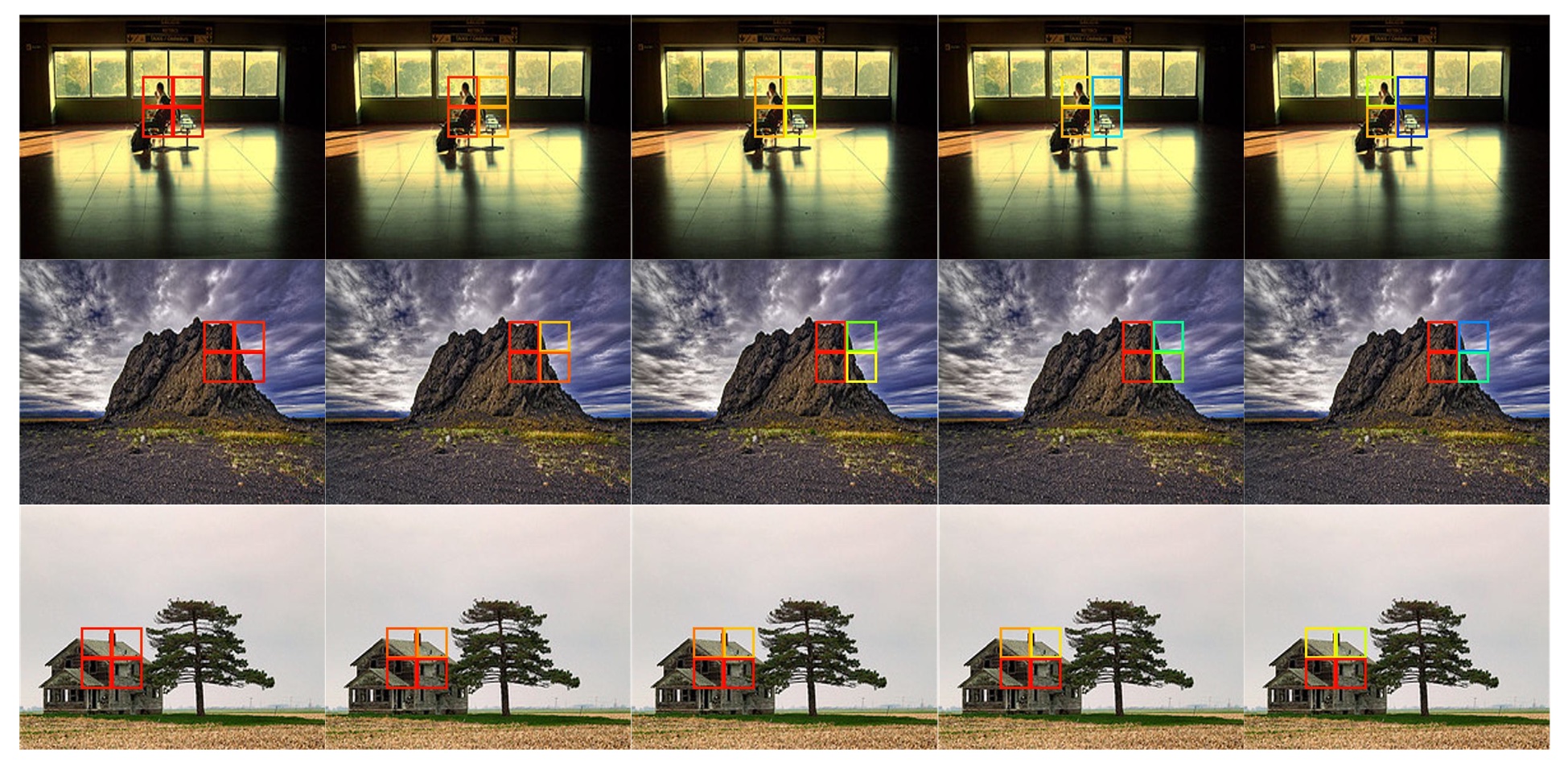}
    \caption{Visualization of evolution of the learned contexts on the MS-COCO dataset. From left to right, the images show the learned importance of cells within a macro-cell at different training iterations: start point, 50th epochs, 100th epochs, 150th epochs and 200th epochs. Redder colors stand for higher importance and bluer are lower.} \label{fig:Visualization_of_evolution_contexts}
\end{figure*}

\subsection{Visualization} \label{sec:visualization}
In this section, we visualize the learned multi-order and cross-scale contexts, as well as the classification results of the instances. Specifically, the network adopts an initial grid of $8\times10$ cells, generating hierarchical features at multiple scales, including $8\times10$, $4\times5$, $2\times3$, $1\times2$ and $1\times1$. At finer scale, the second-order contexts are aggregated through a three-layer context-aware module to enhance local representation. In contrast, at coarser scale, only first-order contexts are aggregated to improve semantic abstraction efficiency.

Firstly, we visualize the learned multi-order context neighborhood on the NUS-WIDE dataset. The image instances with the learned multi-order domain contexts (1st, 3rd and 5th columns) and also their influences on the center cell (2nd, 4th and 6th columns) in different depths from top to bottom are respectively shown in Fig.~\ref{fig:Visualization_of_Numbers_of_Layers} . As the network depth increases, the proposed model can progressively capture more domain structures and focus on more semantically meaningful regions. These results show that deeper networks increasingly highlight a progressive refinement of focus towards meaningful visual regions.

Secondly, we show the influence of each cell in the learned cross-scale context-aware features. As shown in Fig.~\ref{fig:Visualization_of_learning_results}, the 1st, 3rd, and 5th columns show the initial image partitions ($8\times10$, $4\times5$, $2\times3$). The macro-cell in the coarse scale can be visually divided into $2\times2$ micro-cells in the fine scale with overlap. The 2nd, 4th, and 6th columns visualize the impact of different micro-cells within a macro-cell to the learned context-aware features. As mentioned earlier, we first select the most influential micro-cell in each macro-cell as the dominant one, and then perform a weighted fusion of the features from the remaining micro-cells. Qualitative analyses of the visualizations confirm that the proposed network can effectively combine the most discriminative micro-cells within the macro-cell, thereby enhancing feature representation for robust multi-label classification.

Furthermore, we also visualize the evolution procedure of the learned context. Fig.~\ref{fig:Visualization_of_evolution_contexts} shows the visualization of the evolution of the importance of micro-cells within a macro-cell at different training epochs on the MS-COCO dataset. From left to right, the images sequentially show the initial context weights, and the intermediate contexts after 50 epochs, 100 epochs, and 150 epochs as well as the convergence after 200 epochs. The higher the color saturation of the edge, the greater the importance of the corresponding micro-cell. From these images, we observe that the network is able to gradually learn a preference for the visually interesting regions and assign higher weights to the cells \textcolor{black}{as learning evolves}. This dynamic evolution procedure clearly shows how the network learns the structure of the image, focusing on the crucial region of the visual content.

\begin{figure*}[tbp]
    \centering 
    \includegraphics[width=\textwidth]{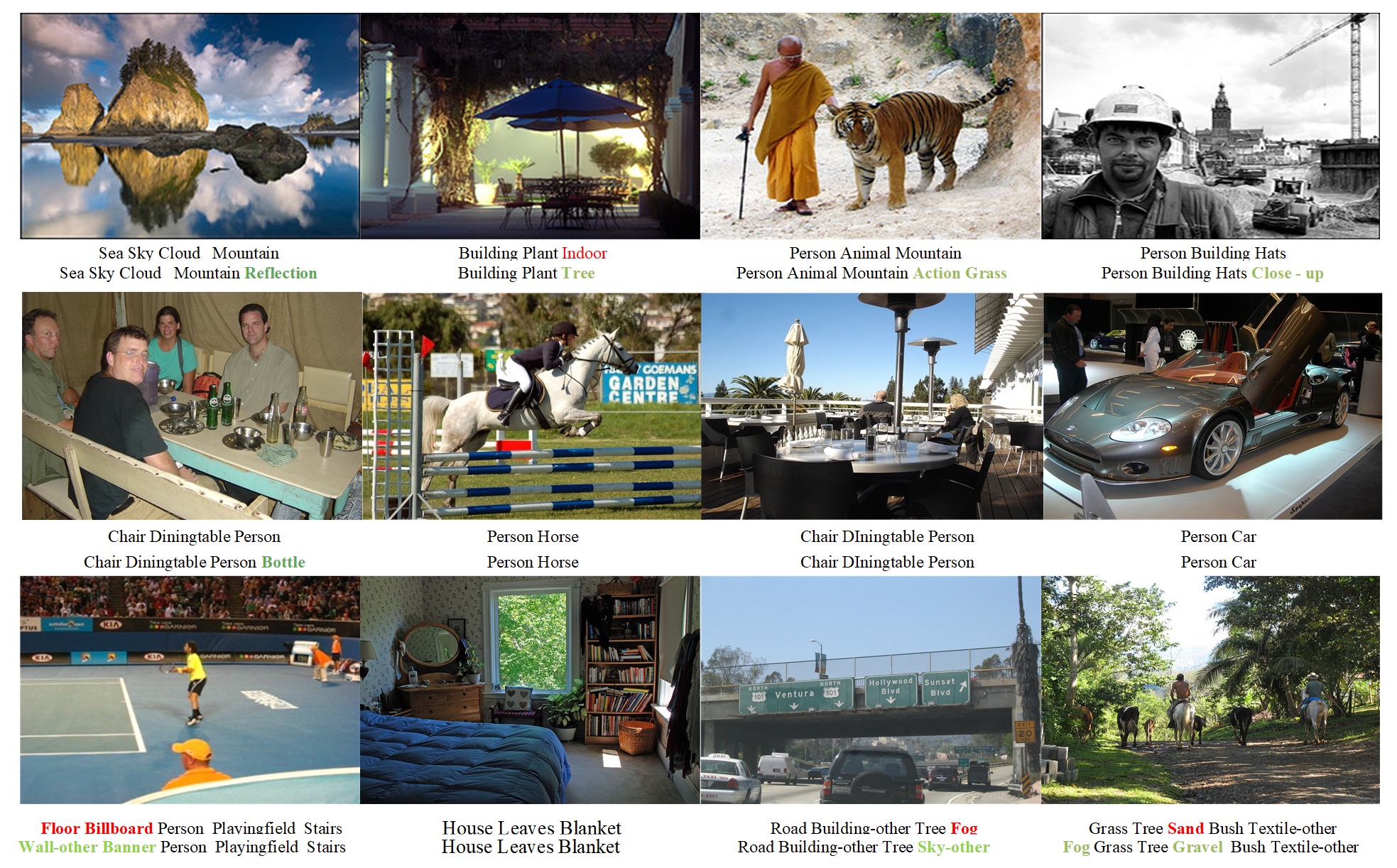}
    \caption{Visualization of classification results of the instances on three benchmark datasets(from top to bottom: NUS-WIDE, PASCAL VOC2007, and MS-COCO). For each image, the classification results are given below where the first row shows the classified labels with a grid of $4\times5$ cells and the second row shows the classified labels with a grid of $8\times10$ cells. The labels in black indicate correct results, the ones in red stands for incorrect ones, and the ones in green highlights newly classified correct labels for $8\times10$ cells. }\label{fig:Annotation_Results}
\end{figure*}

In addition, Fig.~\ref{fig:Annotation_Results} shows the classification results of the proposed networks on the three datasets. It is clearly seen that when the image is divided into a grid of $4\times5$ cells, the learned networks show poor performance in recognizing objects such as ``reflections'', ``bottles'' and ``motorcycles'' in the image, and it is likely to misclassify as other labels with similar features. However, when the initial grids is adjusted to $8\times10$ cells, finer detail features and richer global information allow the models to successfully recognize these challenging objects.

\section{Conclusion} \label{sec:conclusion}

This paper introduces a novel deep cross-scale multi-order context-aware network (dubbed as PanCAN: Panoptic Context Aggregation Networks) for multi-label image classification. This network significantly extends prior context-aware kernel networks by effectively learning multi-order local contexts across multiple scales, leveraging both attention mechanisms and random walks. The integration of global features further yields more discriminative context-aware features. Extensive experiments on NUS-WIDE, PASCAL VOC2007, and MS-COCO demonstrate our network's superior performance over state-of-the-art approaches. Our model exhibits strong capabilities in recognizing objects across diverse scales and complex backgrounds, and its consistent performance across various backbones (ResNet101, TResNetL, and CvT-w24) underscores its robustness to initial feature representation and its efficacy in building enhanced context-aware features.

\newpage

\vfill
\end{document}